\documentclass[journal,comsoc]{IEEEtran}

\usepackage[T1]{fontenc}

\usepackage{ifpdf}
\usepackage{cite}
\usepackage{graphicx}
\graphicspath{{./Fig/}}
\usepackage{amsmath,amsfonts}
\usepackage{bm}
\usepackage{algorithm}
\usepackage{algorithmic}
\usepackage{array}
\usepackage[caption=false,font=normalsize,labelfont=sf,textfont=sf]{subfig}
\usepackage{url}
\begin{document}

\title{Finite Element Modeling of Surface Traveling Wave Friction Driven for Rotary Ultrasonic Motor}

\author{Zhanyue Zhao, Yang Wang, Charles Bales, Yiwei Jiang, Gregory Fischer
\thanks{Zhanyue Zhao, Yang Wang, Charles Bales, Yiwei Jiang, and Gregory Fischer are with the Department of Robotics Engineering, Worcester Polytechnic Institute, Worcester, MA 01605 USA (e-mail: zzhao4@wpi.edu, gfischer@wpi.edu).}
\thanks{This research is supported by National Institute of Health (NIH) under the National Cancer Institute (NCI) under Grant R01CA166379 and R01EB030539.}
}
\maketitle

\begin{abstract}

Finite element modeling (FEM) is a critical tool in the design and analysis of piezoelectric devices, offering detailed numerical simulations that guide various applications. While traditionally applied to eigenfrequency analysis and time-dependent studies for predicting excitation eigenfrequencies and estimating traveling wave amplitudes, FEM's potential extends to more sophisticated tasks. Advanced FEM applications, such as modeling friction-driven dynamic motion and reaction forces, are essential for accurately simulating the complex behaviors of piezoelectric actuators under real-world conditions. This paper presents a comprehensive motor model that encompasses the coupling dynamics between the stator and rotor in a piezoelectric ultrasonic motor (USM). Utilizing contact theory, the model simulates the complex conditions encountered during the USM's initial start-up phase and its transition to steady-state operation. Implemented in COMSOL Multiphysics, the model provides an in-depth analysis of a rotary piezoelectric actuator, capturing the dynamic interactions and reaction forces that influence its performance. The introduction of this FEM-based model represents a significant advancement in the simulation and understanding of piezoelectric actuators. By offering a more complete picture of the motor's behavior from start-up to steady state, this study enables more accurate control and optimization of piezoelectric devices, enhancing their efficiency and reliability in practical applications.

\end{abstract}

\begin{IEEEkeywords}

Finite Element Modeling, Surface Traveling Wave, Ultrasonic Motor, Friction Driven Simulation

\end{IEEEkeywords}

\section{Introduction}

\IEEEPARstart{A} huge amount and effort of work were applied in the USM simulation and FEM simulation from different groups, however, most of the simulations were focused on the excitation of stator only \cite{carvalhomultiphysics,patel2012design,zheng2015effects, liu2019design,li2017traveling,yang2020design,li2022dynamics, chen2006traveling,carvalho2020study,zhao2021preliminary,zhao2023preliminary,zhao2024study,zhao2024characterization}. Take COMSOL Multiphysics as an example, the stator assembly simulation and analysis are easy to process with the multiphysics node called a piezoelectric device. Current stator simulation work only finishes half of the motor modeling, the other half will be the coupling modeling between rotor and stator, and very limited research and studies achieve the whole model of stator and rotor. Frangi \emph{et al.} developed a FEM simulation for rotating piezoelectric ultrasonic motor using Abaqus/Explicit\textregistered~(Dassault Systemes, France)\cite{frangi2005finite}, and their model was an optimized code for highly non-linear analyses involving contact and dynamic effects. Ren \emph{et al.} presented a FEM model for traveling wave rotary ultrasonic motor (TRUM) using ADINA\textregistered~(Watertown, MA, USA) \cite{ren2019output}, and the 3D FE model simulating the output performance simulation and contact analysis of TRUM was in good agreement with the available experimental data. Basudeba \emph{et al.} developed a complete FEM simulation for a dual friction-drive (DFD) surface acoustic wave (SAW) linear motor using COMSOL Multiphysics \cite{behera2019investigating}. The friction mechanism in tribology indicates that the contact mechanism of USM is a greatly complex non-linear problem, which makes it difficult to develop an analytical or semi-analytical model. Building up a FEM simulation using COMSOL Multiphysics will provide a powerful tool for the structural design and performance prediction of new USMs design, and parameters testing, and facilitate the stress analyses to avoid the failure of the motor.

\section{Finite Element Modeling of A USR30 Motor}

\begin{figure}
    \centering
    \includegraphics[width=0.9\linewidth]{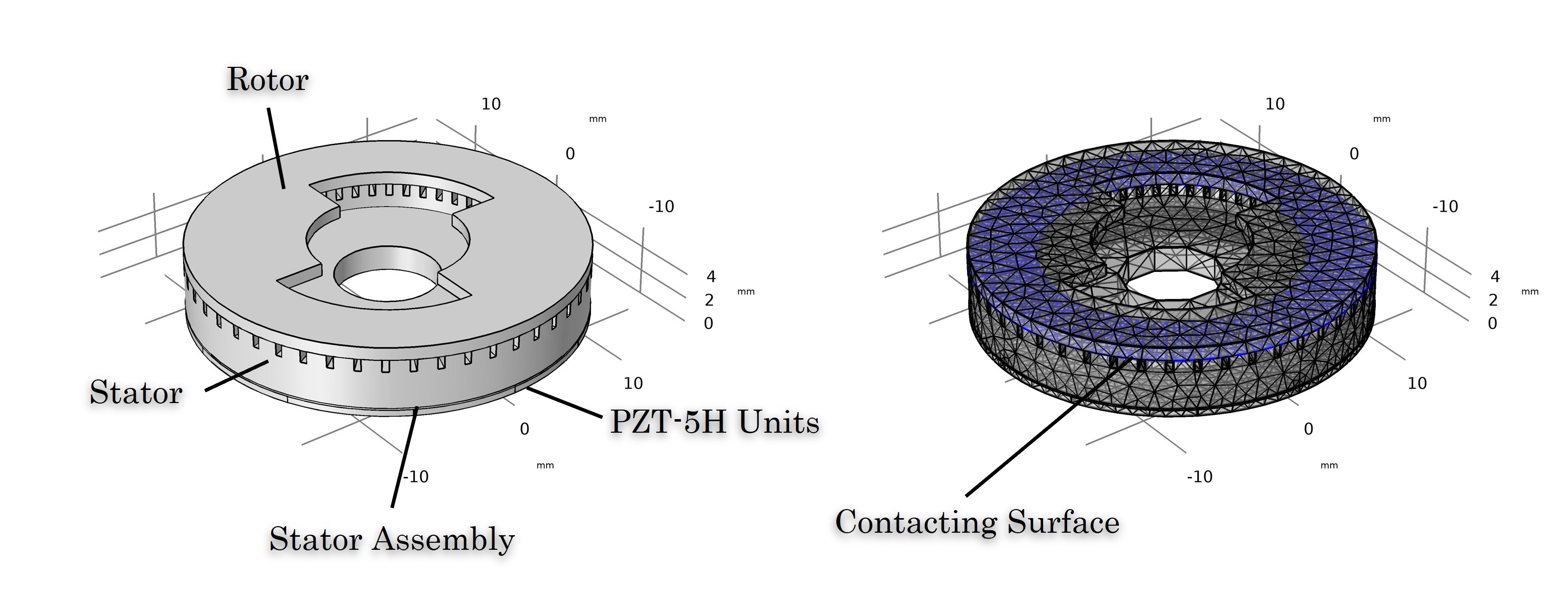}
    \caption{(Left) Modeling geometry and component definition. (Right) Component mesh with extra fine on the contacting surface, and normal on the rest of the component.}
    \label{fig:mesh}
\end{figure}

A FEM simulation based on the USR30 motor using COMSOL Multiphysics was developed in this section. Based on our previous work \cite{carvalho2020study,zhao2021preliminary,zhao2023preliminary,zhao2024study}, a rotor geometry was added to the current component. A contact pair was created in the definitions node so that the stator assembly and rotor geometry were able to create pairs in the form assembly mode. Note that to visually observe the rotary motion, 2 symmetric fins towards the rotor center were added, and these cutouts should have a negligible effect on the performance. The geometry can be found in Figure \ref{fig:mesh} left figure. For initial modeling, the rotor was defined as aluminum while the stator was defined as copper from the library, and the material properties of components are shown in Table \ref{tbl:properties}. The piezoelectric material properties are shown in Table \ref{tbl:pztcoupling}, \ref{tbl:relative}, and \ref{tbl:elasticity}. In the solid mechanics physics node, a contact node was added with the contact pair defined previously. The bottom of rotor and the all the top surfaces of teeth from the stator were selected in the pair, where the source boundary was the rotor and the destination boundaries were stator teeth. A rotary constraint was also added onto the rotor to force it to rotate along the z-axis, which was the center shaft held and constrained by bearing in the real world. A 50N load was applied equally on the top surface of the stator acting as prepressure, and the coefficient of friction (COF) $\mu$ was defined as 0.2, which is the optimized value reported in \cite{van2011dynamic}. In the mesh step, the contact surface on both rotor and stator was defined as the extra-fine tetrahedral size, and the other surface and volume were defined as normal to speed up the calculation. A meshed image is shown in Figure \ref{fig:mesh} right figure. An eigenfrequency and a time-dependent study were created, where the eigenfrequency was used to find the proper eigenfrequency (we used $4^{th}$ mode value as the main study excitation frequency), and we defined 0 to 5ms with 0.01ms incremental range in the time-dependent to study complex motion performance. Note that to avoid the large memory occupation issue, the Pardiso solver was selected to replace the MUMPS solver. 

\begin{table}
    \caption{Basic mechanical properties of plastic stator components.}
    \vspace{1mm}
    \centering
    \resizebox{\linewidth}{!}{
    \begin{tabular}{cccc}
    \hline
    Material   & Density {[}$Kg/m^3${]} & Possion's Ratio & Young's Modulus {[}\emph{GPa}{]} \\ \hline\hline
    Ultem 1000 & 1270                & 0.3             & 3.2                       \\ \hline
    Epoxy      & 3500                & 0.33            & 0.7                       \\ \hline
    PZT-5H     & 7500                & -             & -                       \\ \hline
    Copper     & 8960                & 0.35            & 110                      \\ \hline
    Aluminum   & 2700                & 0.33            & 70                       \\ \hline
    \end{tabular}
}
    \label{tbl:properties}
\end{table}

\begin{table}[ht!]
\caption{PZT-5H Coupling Matrix in [$C/m^2$]}
    \vspace{1mm}
    \centering
    \begin{tabular}{|c|c|c|c|c|c|}
    \hline
    0        & 0        & 0       & 0       & 17.0345 & 0 \\ \hline
    0        & 0        & 0       & 17.0345 & 0       & 0 \\ \hline
    -6.62281 & -6.62281 & 23.2403 & 0       & 0       & 0 \\ \hline
    \end{tabular}
\label{tbl:pztcoupling}
\end{table}

\begin{table}[ht!]
\caption{PZT-5H relative
permittivity matrix [1/s]}
    \vspace{1mm}
    \centering
    \begin{tabular}{|c|c|c|}
    \hline
    1704.4 & 0      & 0      \\ \hline
    0      & 1704.4 & 0      \\ \hline
    0      & 0      & 1433.6 \\ \hline
    \end{tabular}
    \label{tbl:relative}
\end{table}

\begin{table}
\caption{PZT-5H Elasticity Matrix in Pascal [\emph{Pa}]. Matrix is symmetric about its diagonal.}
    \vspace{1mm}
    \centering
    \resizebox{\linewidth}{!}{
    \begin{tabular}{|c|c|c|c|c|c|}
    \hline
    1.27205e11 & 8.02122e10 & 8.46702e10 & 0            & 0            & 0            \\ \hline
    -            & 1.27205e11 & 8.46702e10 & 0            & 0            & 0            \\ \hline
    -            & -            & 1.17436e11 & 0            & 0            & 0            \\ \hline
    -            & -            & -            & 2.29885e10 & 0            & 0            \\ \hline
    -            & -            & -            & -            & 2.29885e10 & 0            \\ \hline
    -            & -            & -            & -            & -            & 2.34742e10 \\ \hline
    \end{tabular}
}
    \label{tbl:elasticity}
\end{table}

\begin{figure}
    \centering
    \includegraphics[width=0.5\linewidth]{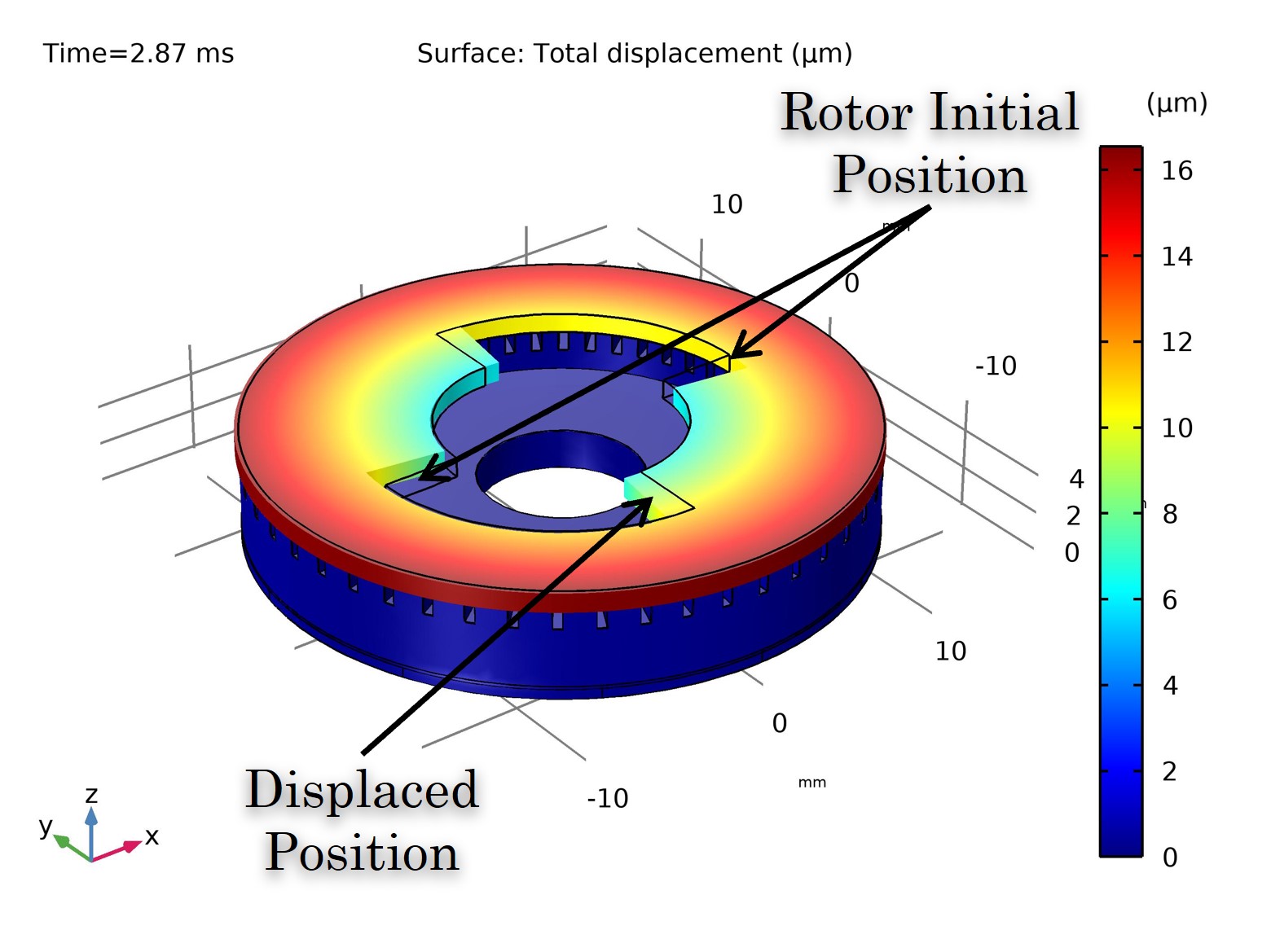}
    \caption{Displacement profile of the rotor current position compared to the initial position.}
    \label{fig:move}
\end{figure}

\begin{figure}
    \centering
    \includegraphics[width=0.9\linewidth]{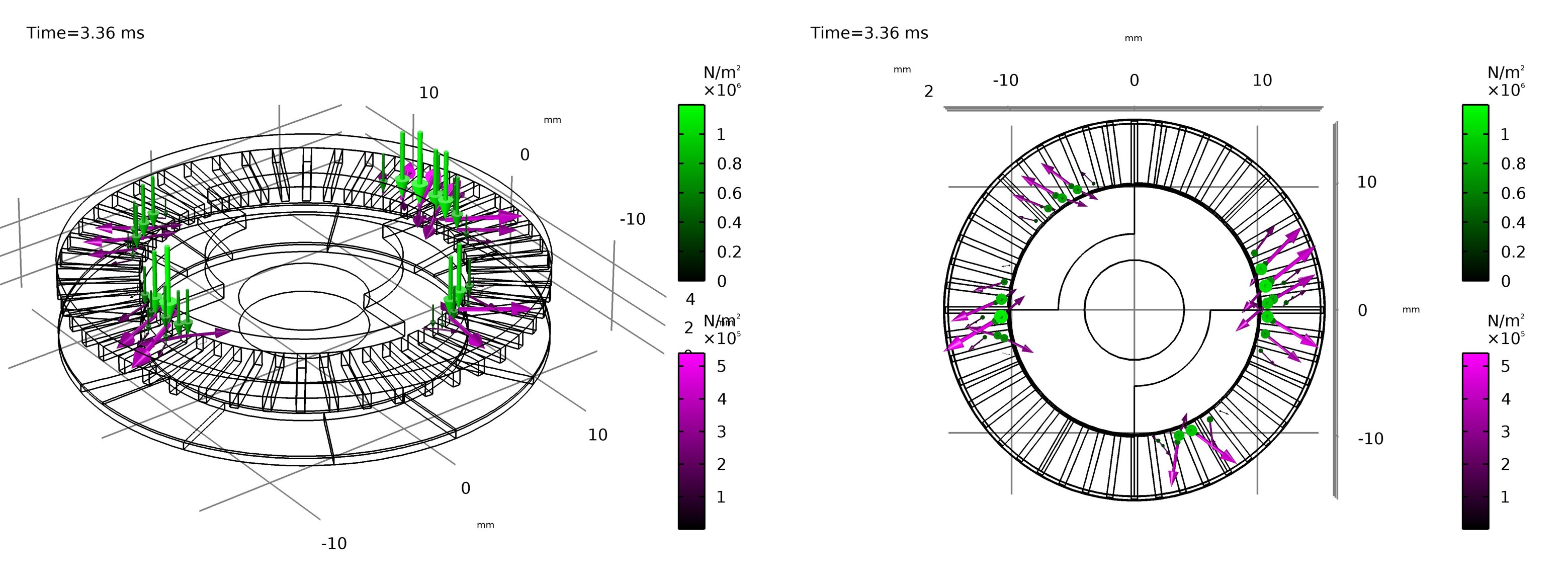}
    \caption{Normal reaction force (green) and friction force (purple) intensity. The simulation was performed under 4$^{th}$ mode so 4 peaks are shown in the figure. The friction was tangential along the rotor circumference to drive it to rotate.}
    \label{fig:force}
\end{figure}

\begin{figure}
    \centering
    \includegraphics[width=1\linewidth]{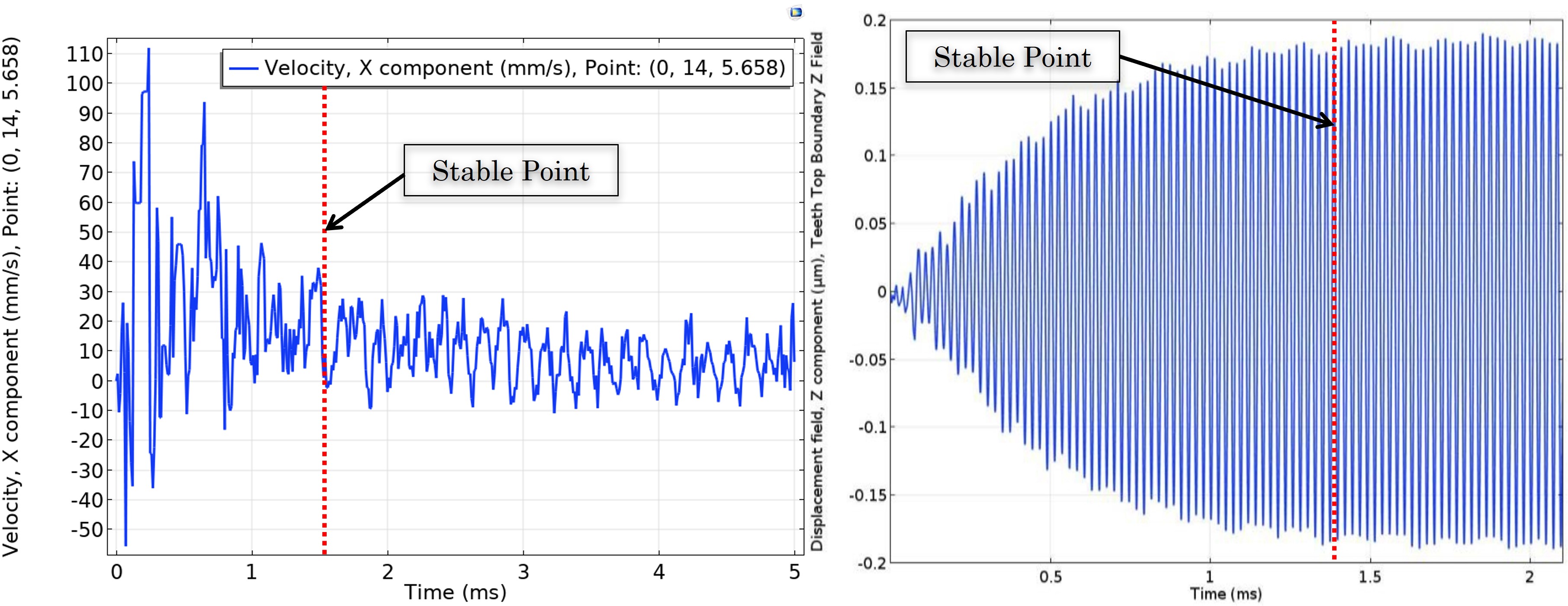}
    \caption{(Left) Results of an x-component velocity probe placed on the rotor. Velocity reaches a settle-down status at approximately 1.5ms. (Right) Carvalho \textit{et al.} developed a stator simulation with the z-axis displacement reaching a settle-down status of around 1.4ms. Image reproduced from \cite{carvalhomultiphysics}.}
    \label{fig:stable}
\end{figure}

\begin{figure}
    \centering
    \includegraphics[width=0.7\linewidth]{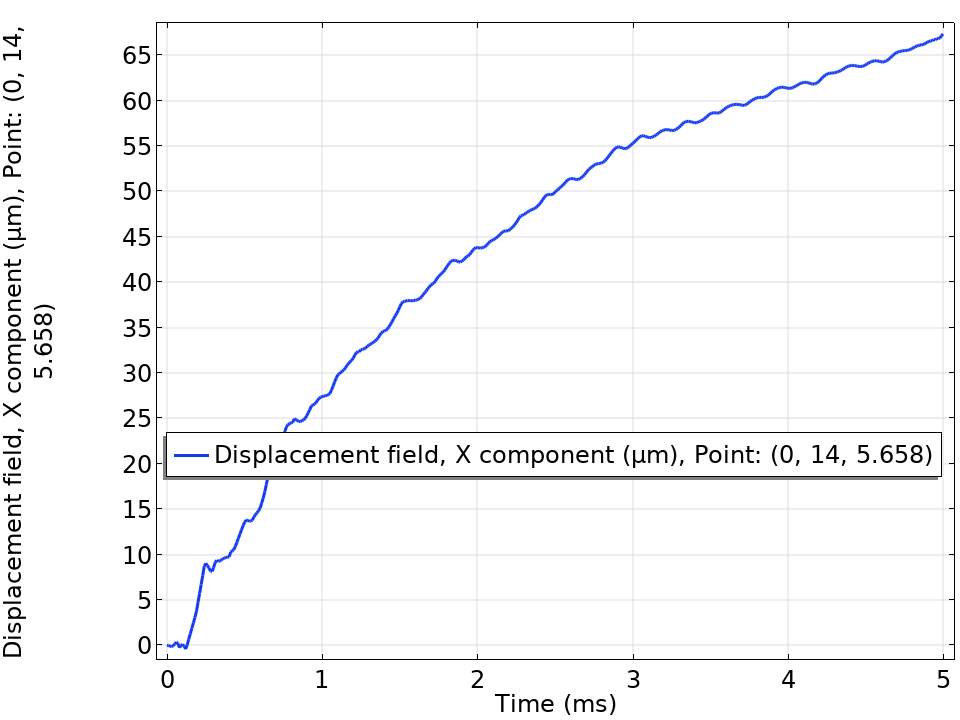}
    \caption{Result of an x-component displacement probe placed on the rotor. It reaches over 65$\mu m$ at the end of the simulation.}
    \label{fig:disp}
\end{figure}

\begin{figure}
    \centering
    \includegraphics[width=0.7\linewidth]{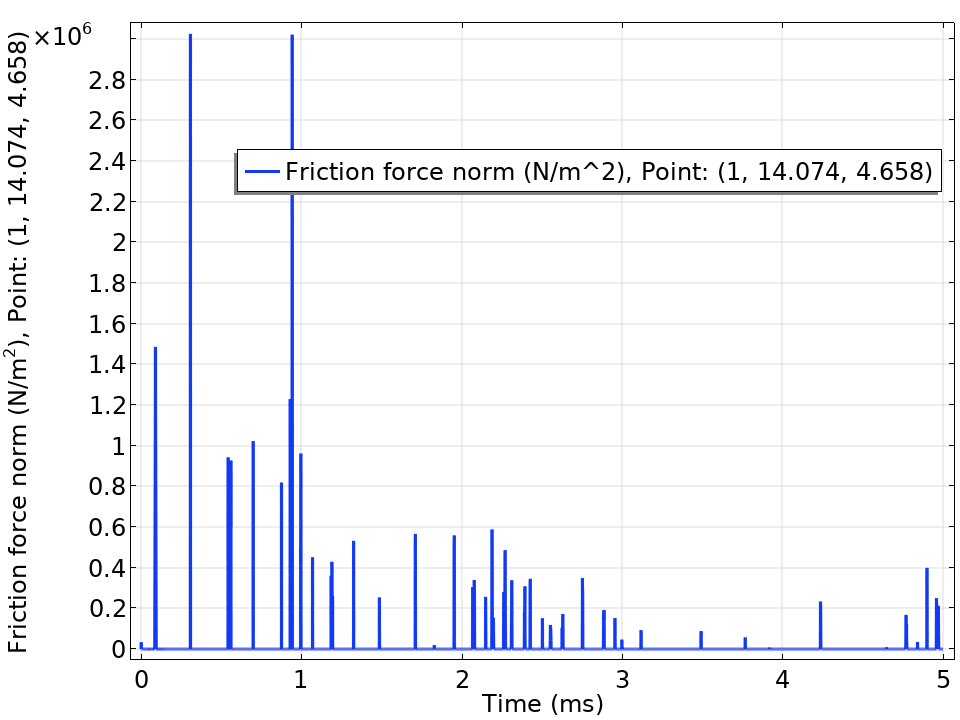}
    \caption{Result of friction force from the probe location.}
    \label{fig:forcenormal}
\end{figure}

\begin{figure}
    \centering
    \includegraphics[width=0.8\linewidth]{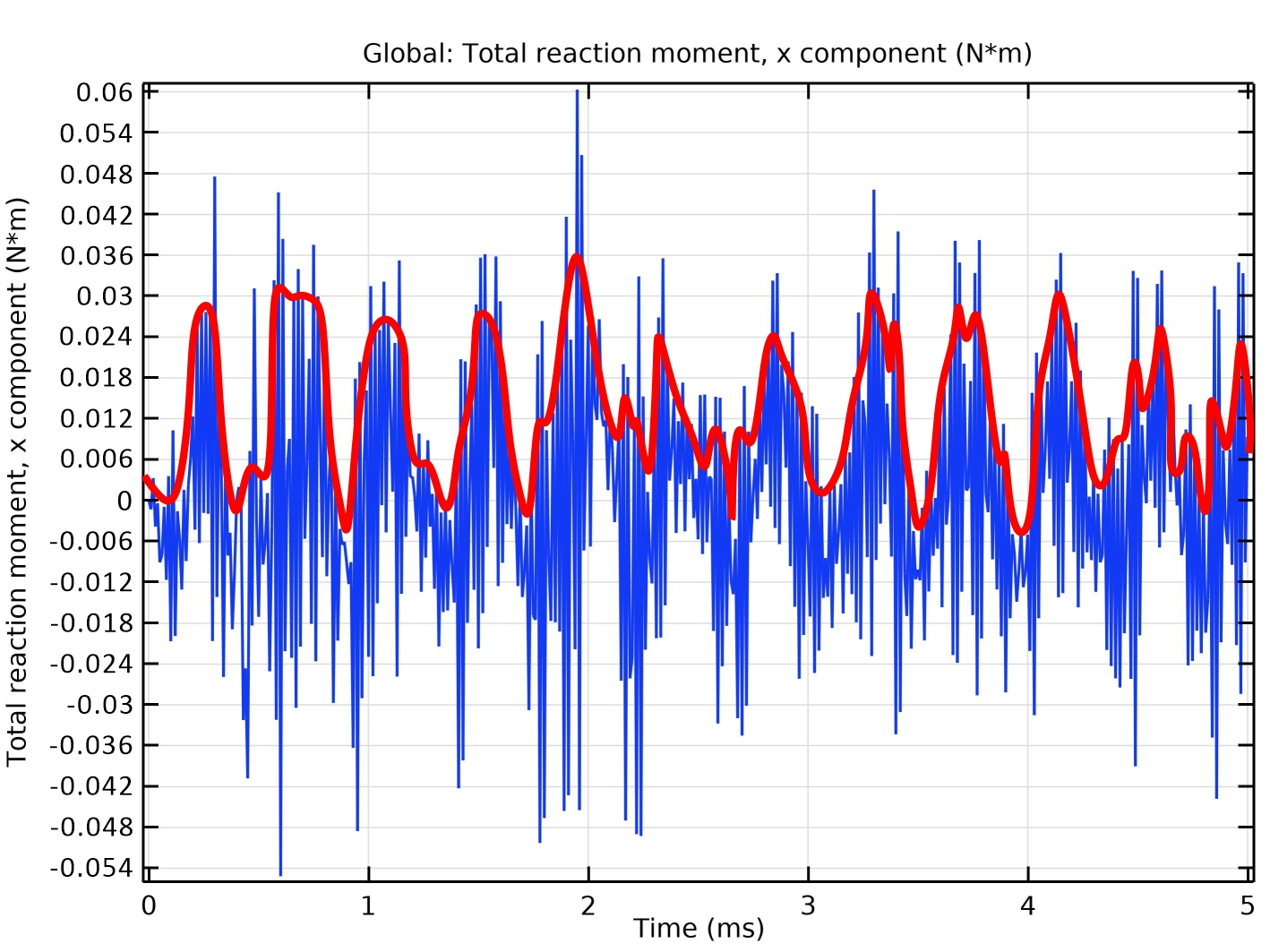}
    \caption{Result of an x-component total reaction moment probe, which shows the periodical reaction torque transmitted to the rotor. We considered the reaction torque value after 1.5ms (red dashed line) when the motor reached a steady state, and read the average value (orange dashed line) of torque enveloping data (red solid curve).}
    \label{fig:torque1}
\end{figure}

Figure \ref{fig:move} represents the total displacement 3D results, and it clearly shows the displacement of the rotor from an initial position (black line frame) to a displaced position (colored solid structure) at 2.87ms. Multiple probes were placed at the rotor contacting surface, namely velocity, displacement, friction force, and reaction moment (torque). Figure \ref{fig:force} shows the pressure (green arrow) and friction force (green arrow) density and direction map. The four peaks of pressure matched the 4$^{th}$ excitation mode we configured, and the friction force performed the tangential direction portion, which indicates the driving force applied onto the rotor and pushed it to spin. An x-component velocity probe was placed on the rotor and the results can be found in Figure \ref{fig:stable} left figure. The result shows that the rotor speed was settled down with stable amplitude output at around 1.5ms, which matches the results reported by Carvalho \textit{et al.} with the z-axis displacement reaching a settle down status around 1.4ms \cite{carvalhomultiphysics}. Figure \ref{fig:disp} shows the displacement probe, which indicates the rotor was moving, at the end of the model and reached over 65$\mu m$ along the x-axis. Figure \ref{fig:forcenormal} shows the friction force from the probe location, and at the beginning of spinning the friction interfered with the rotor, the friction was large, and when it reached a steady state the friction was performed in a balanced range periodically in one point. A detailed analysis will be discussed in a later section. Figure \ref{fig:torque1} shows the total reaction moment in the x-component, and the results indicate that periodical torque was transmitted to the rotor when the motor reached a steady state. Here we only considered the reaction torque value after 1.5ms (red dashed line) when the motor reached a steady state, and we read the average value (orange dashed line) of torque enveloping data (red solid curve). Note that in the current stage, this is a simplistic interpretation of the motor torque, and future studies of how the periodic torque correlates to the output torque of the motor are required. 

\section{Prepressure Modeling and Experimental Validation}

The prepressure or pre-load applied on the rotor has a significant effect on USM performance since it is mainly driven by friction on the contacting surface. Many studies have been put in for standard USM prepressure effect, especially studies on Shinsei USR30 and 60 series motors. However, studies on prepressure were mostly using commercial motors, specifically commercial stators, very few studies focus on changing the material of the motor for MRI compatibility consideration, so studies for new material stators with parameters optimization are still blank in the community. In this section, a model focusing on the prepressure effect of the USM will be presented, and then validated with an experiment. 

\begin{figure}
    \centering
    \includegraphics[width=0.8\linewidth]{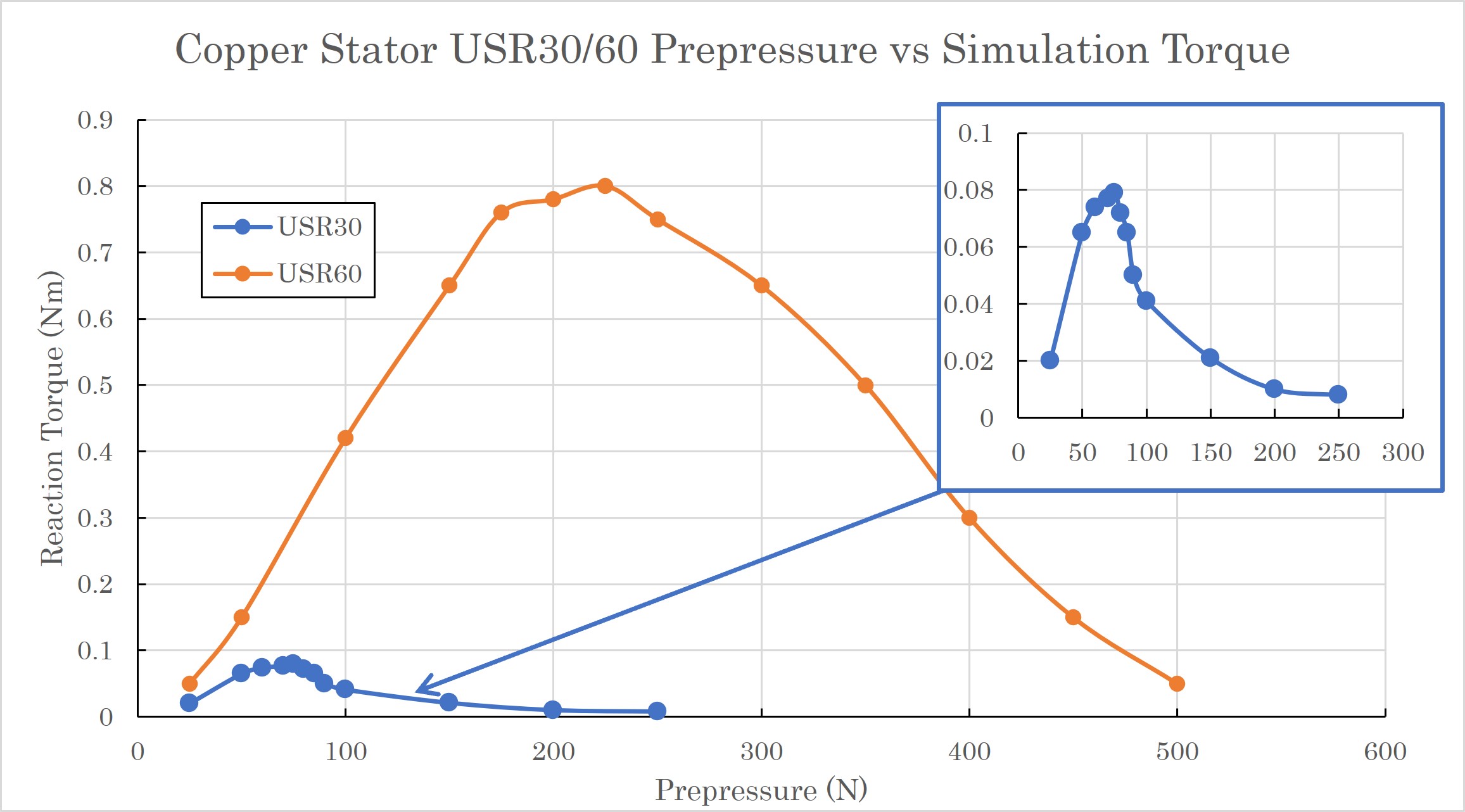}
    \caption{Reaction torque value when the motor became a steady state under different prepressure conditions.}
    \label{fig:preloadsim}
\end{figure}

To evaluate our model, we first used a copper stator as the target baseline. A USR30 and a USR60 motor geometry were built using the default 3D modeling tool in COMSOL Multiphysics, then define the stator assembly and rotor as copper, epoxy, PZT-5H, and aluminum respectively, material properties can be found in our previous work \cite{carvalho2020study,zhao2021preliminary}. In the solid mechanics physics node, we already added a boundary load on the top surface of the rotor. In this model, we used the auxiliary sweep function in study extensions under the study setting class to put in from 25 to 250N and from 25 to 500N with 25N incremental value with USR30 and USR60 models respectively. In this model, the probe was placed on the stator, and the reaction torque result from Figure \ref{fig:torque1} yields a periodic changing value. Based on Figure \ref{fig:stable}, the motor reached a steady state at approximately 1.5ms, here we read the average value of torque enveloping data (neglecting the spike data) after 1.5ms as the reaction torque value. We recorded the reaction torque value when the motor became a steady state under different prepressure conditions, and the result is shown in Figure \ref{fig:preloadsim}. From the results, we can know the highest torque performed up to 0.8Nm for USR60 at an estimate of 225N prepressure, while 0.08Nm for USR30 at an estimate of 75N prepressure.

\begin{figure}
    \centering
    \includegraphics[width=0.9\linewidth]{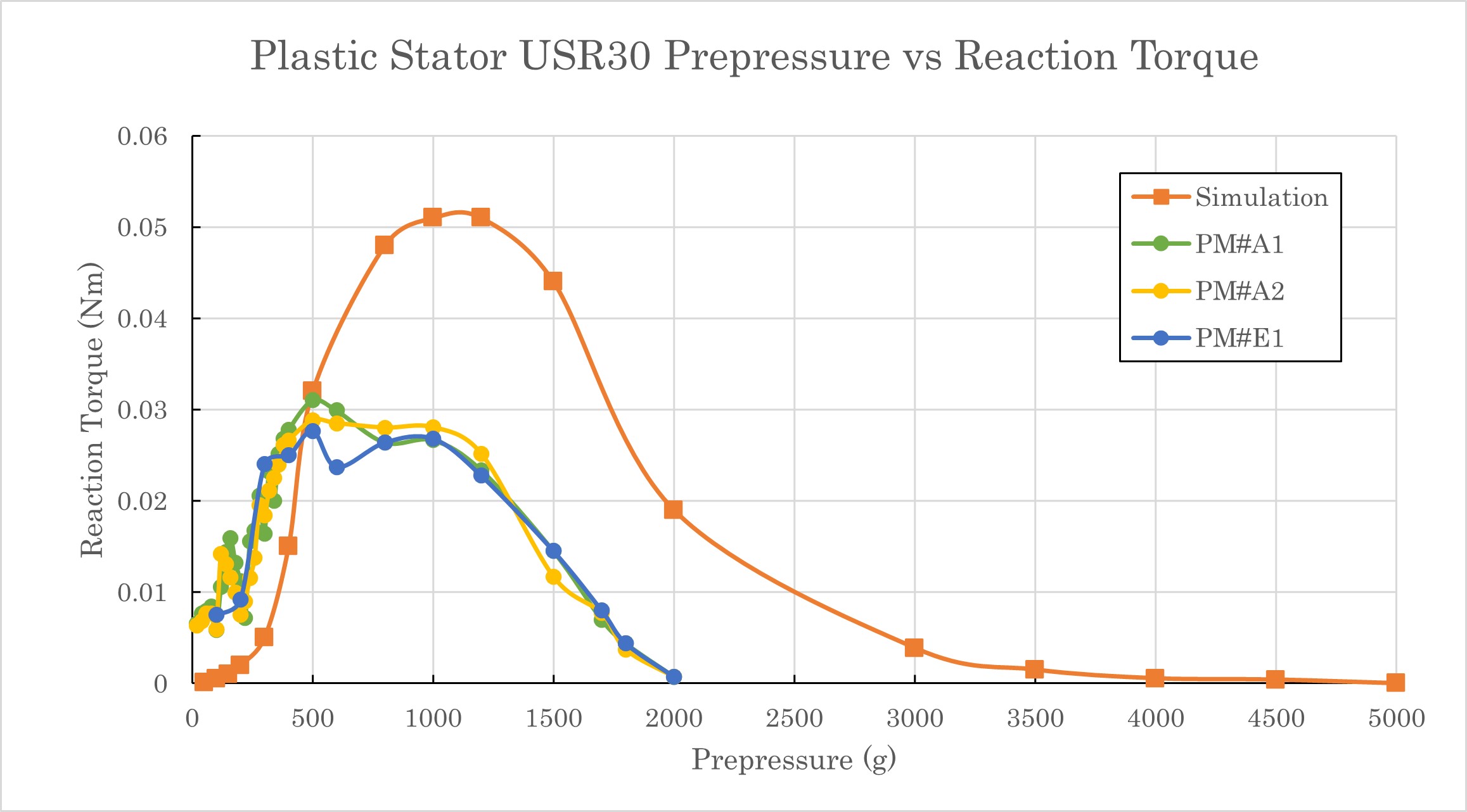}
    \caption{The reaction torque data from simulation and the experimental results from A1, A2, and E1 three equivalent plastic motors output torque performance.}
    \label{fig:load_simu_exp}
\end{figure}

Based on the model of copper stater USR30 model, we switched the stator material to Ultem plastic. In this model, we changed the prepressure unit from Newton to Gram to match our experimental results from our previous work \cite{zhao2024design}, and the value we configured in the auxiliary sweep was from 20 to 5000g. The results of the reaction torque and the experimental results from A1, A2, and E1 three equivalent plastic motors output torque performances are shown in Figure \ref{fig:load_simu_exp}. It was observed that in the simulation the highest reaction torque reached up to 0.052Nm performed at approximately 1000-1200g range prepressure, versus the experiment results of the highest torque up to 0.032Nm performed at 500-1000g range prepressure. There is an approximate 40\% overshoot in torque performance between the simulation and experimental result and a 400g prepressure data shift, the following reasons may cause this issue. First, these motors were all fabricated manually, which may contain dimensional error and tolerance, these might affect the motor performance compared to a simulation environment of idealized state and condition. The other reason may be caused by the surface roughness and coefficient of friction ($\mu$) misalignment issue, where we were using 0.2 with $\mu$ value in simulation versus using grit number 1000 sandpaper finished surface with experimental. Furthermore, the parameters and configuration used in the current stage simulation may still not be accurate and appropriate.

\section{Coefficient of Friction Modeling and Surface Roughness Validation}

Besides prepressure, the coefficient of friction (COF) also plays a significant effect on USM performance, because the classical laws of friction from Coulomb's model suggest that friction force is proportional to the normal load and COF. In this section, a study of COF and surface roughness effect on the motor torque is presented. 

The COF ($\mu$) is a dimensionless scalar that equals the ratio of the force of friction between two bodies and the force pressing them together, either during or at the onset of slipping with a range near zero to greater than 1. The coefficient of friction mainly depends on the materials used, also related to the surface roughness to some degree. Much work has been conducted to explore the relationship between COF and surface roughness, which under some specific conditions are proportional to each other and affect the friction force to the same degree \cite{feng2017surface,larsson2009modelling}. In this study, we used a brass stator as an initial modeling test as a baseline similar to prepressure simulation. Considering the Teflon tribology layer was used in the real USR30 motor, in our brass model the rotor was switched from aluminum to polytetrafluoroethylene (PTFE), or Teflon. Because the rotor was defined as a rigid body the deformation of soft material can be ignored, otherwise the Teflon rotor will be fully attached to the stator along the wave pattern, then no rotary motion will perform. Note that considering the Ultem is soft compared to brass and the COF between Ultem and metal is higher than PTFE and brass, we keep the aluminum rotor in Ultem plastic simulation. This configuration is based on the surface acoustic wave (SAW) driven motor modeling reported by Behera \textit{et al.} in \cite{behera2019investigating}, and this was considered a very small contact area with multiple wave amplitude peaks. A future study of configuring the rotor as a non-rigid body is needed.

\begin{figure}
    \centering
    \includegraphics[width=0.9\linewidth]{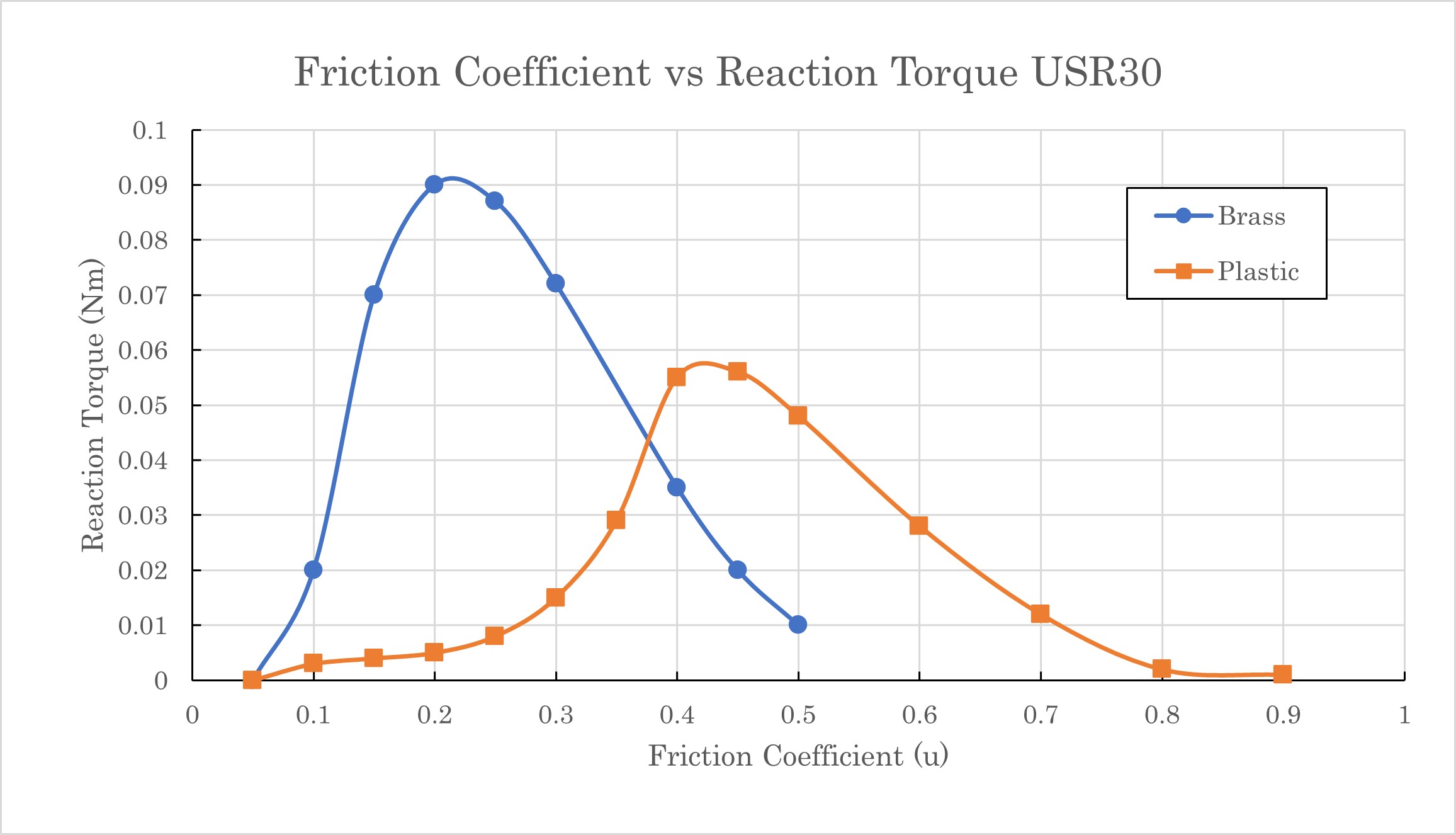}
    \caption{Simulation result of reaction torque result under different COF conditions between stator and rotor.}
    \label{fig:FCsim}
\end{figure}

Figure \ref{fig:FCsim} shows the simulation result of reaction torque result under different COF conditions between stator and rotor, where the brass stator reached up to 0.09Nm reaction torque under COF=0.2 and plastic stator reached up to 0.055Nm torque under approximate COF=0.4. The COF between Teflon and brass yields a range of 0.05 to 0.2, and laws of friction suggest the higher COF generates larger friction. Also based on the report in \cite{zhao2011ultrasonic,sharp2010design} the optimized COF suggests the range of 0.15 to 0.3 in traditional USM design. The simulation results of the brass stator matched the current studies and material properties, namely at approximately $\mu = 0.2$, which the motor will reach the highest torque of 0.09Nm.

\begin{figure}
    \centering
    \includegraphics[width=0.9\linewidth]{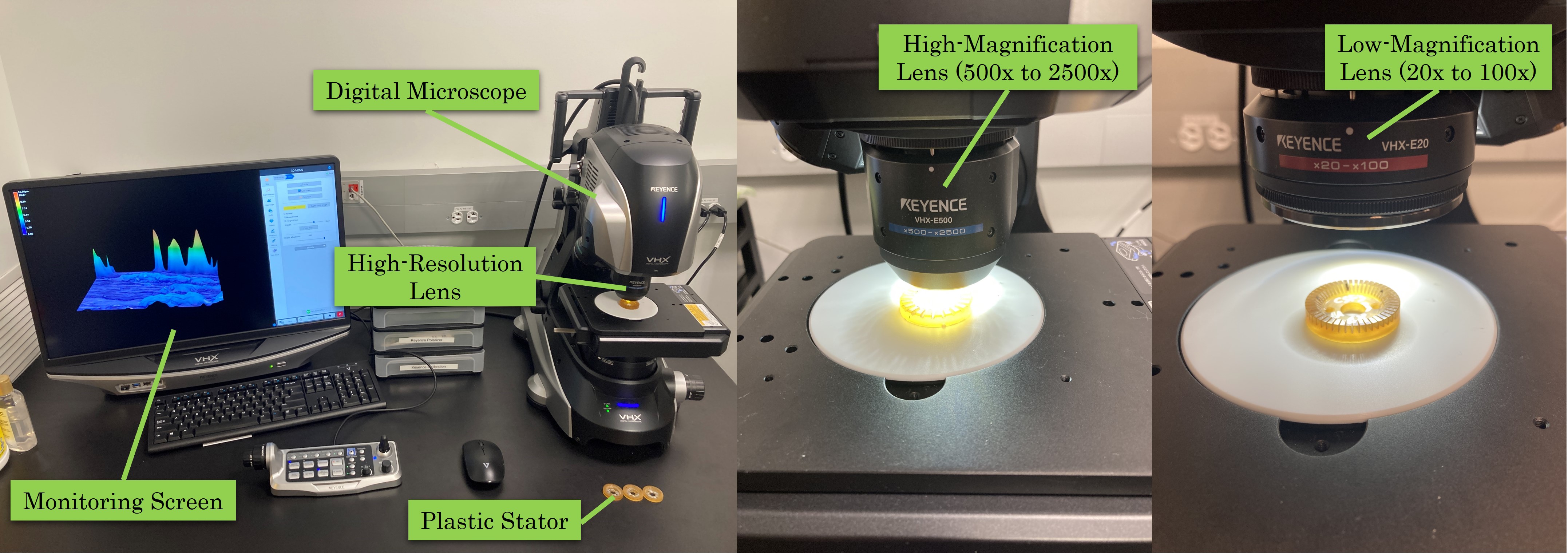}
    \caption{Setup of the microscope.}
    \label{fig:setup1}
\end{figure}

\begin{figure}
    \centering
    \includegraphics[width=0.9\linewidth]{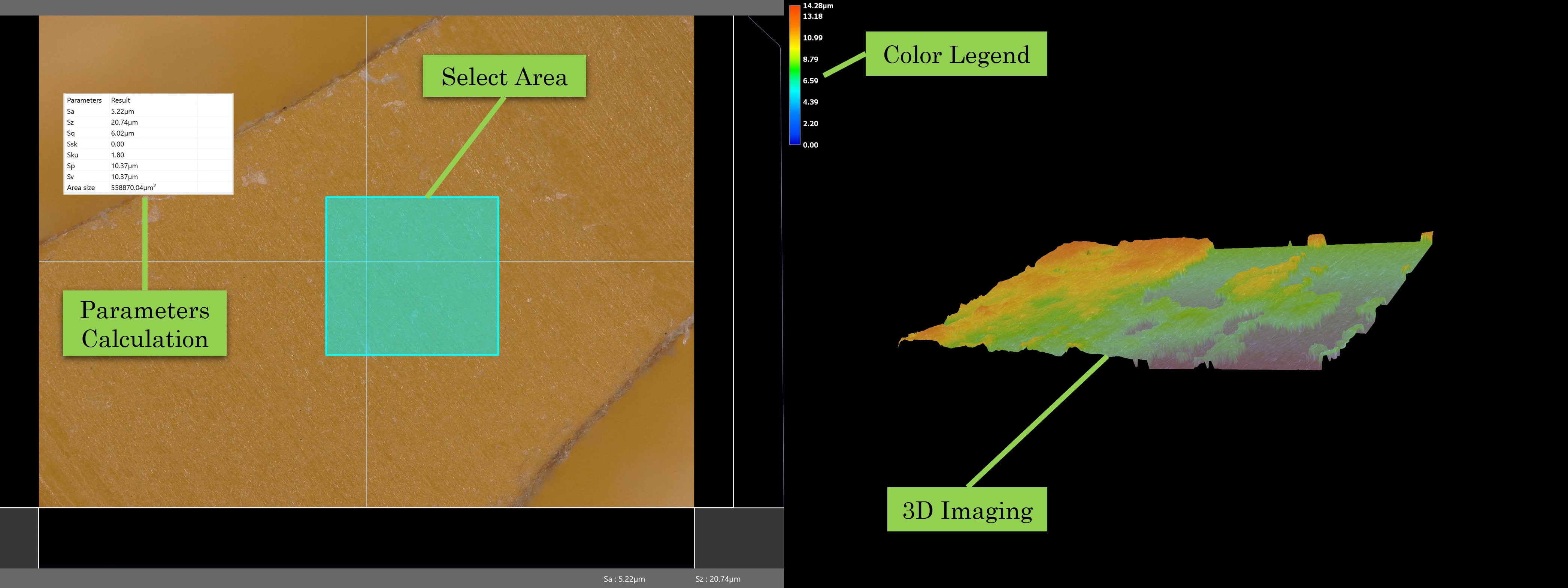}
    \caption{Software level operation process and rendered 3D image generated based on the measurement data.}
    \label{fig:setup2}
\end{figure}

The surface roughness is a component of surface finish (surface texture). It is quantified by the deviations in the direction of the normal vector of a real surface from its ideal form and can be measured by microscope. In tribology, rough surfaces usually have higher friction coefficients than smooth surfaces, and further affect friction. To evaluate the surface roughness created by different grit numbers of sandpaper, a digital optical microscope (VHX-7000N, Keyence, Japan) was used. Figure \ref{fig:setup1} shows the setup of the microscope, which consisted of the digital microscope, various high-resolution lenses, and a monitoring screen. After placing the stator on the stage of the microscope, select the proper lens, then move the lens to the focus location until a clear surface appears on the screen. Figure \ref{fig:setup2} shows the software-level operation process. After focusing on the stator, an area was selected on the contacting surface of stator teeth, and parameters were calculated automatically. A rendered 3D image was also generated based on the selected area measurement data. A list of surface parameters generated by the microscope is shown in Table \ref{tbl:SRpara}, and in this study, we were using the $S_a$ only as the surface roughness indicator.

\begin{table}[ht!]
    \caption{Area Roughness Parameters.}
    \centering
    \resizebox{\linewidth}{!}{
    \begin{tabular}{llll}
    \hline
    Parameters & Name                     & Expression & Unit \\ \hline\hline
    $S_a$      & Arithmetical mean height & $S_{a}=\frac{1}{A}\iint\left|Z(x,y)\right|\,dxdy\ $ & $\mu m$  \\ \hline
    $S_z$      & Maximum height           & $S_{z}=S_{p}-S_{v}$ & $\mu m$   \\ \hline
    $S_q$      & Root mean square height  & $S_{q}=\sqrt{\frac{1}{A}\iint Z^2(x,y)\,dxdy}$  & $\mu m$   \\ \hline
    $S_{sk}$   & Skewness                 & $S_{sk}=\frac{1}{S_{q}^3}\left [ \frac{1}{A}\iint Z^3(x,y)\,dxdy\right ]$  & -  \\ \hline
    $S_{ku}$  & Kurtosis                 & $S_{ku}=\frac{1}{S_{q}^4}\left [ \frac{1}{A}\iint Z^4(x,y)\,dxdy\right ]$  & -  \\ \hline
    $S_p$      & Maximum peak height      & $S_{p}=maxZ(x,y)$  & $\mu m$   \\ \hline
    $S_v$      & Maximum pit height       & $S_{v}=\left|minZ(x,y)\right|$  & $\mu m$   \\ \hline
    Area Size  & Area Size                & -          & $\mu m^2$  \\ \hline
    \end{tabular}
    }
    \label{tbl:SRpara}
\end{table}

\begin{figure}
    \centering
    \includegraphics[width=0.9\linewidth]{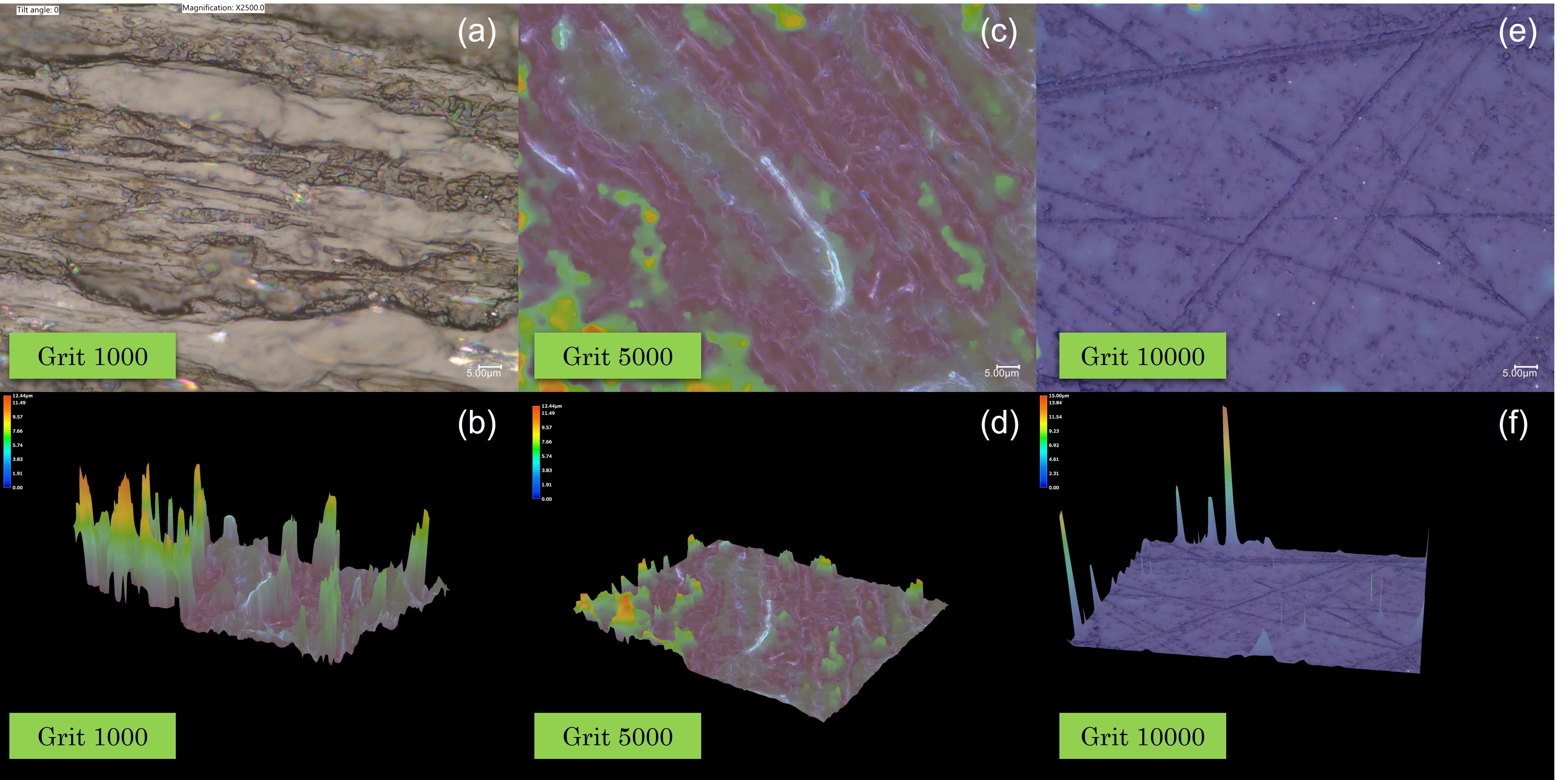}
    \caption{Micro images and 3D rendering images from the surface finish with grit numbers of 1000, 5000, and 10000 respectively.}
    \label{fig:microimg}
\end{figure}

Four stators with different grit numbers of sandpaper finishing were measured, and five sample areas were selected equally along the circumference of the contacting surface on the stator teeth. The mean value of $S_a$ from each stator's five trials measurement data was calculated, and detailed properties of surface roughness measurement can be found in Table \ref{tbl:Sa}. Note that the area size was smaller when measuring the higher grit number finish surface, this is because we were using a high-magnification lens (500 to 2500$\times$) on 5000 and 10000 grit stators compared to using a low-magnification lens (20 to 100$\times$) on 100 and 1000 grit stators. The micro image and 3D rendering images are shown in Figure \ref{fig:microimg}, in which larger grit number sandpaper, or finer sandpaper treated surface finish performed smaller spikes and homogeneous surface finish.

\begin{table}[]
    \centering
    \caption{Surface Roughness Measurement Results}
    \begin{tabular}{|c|c|c|}
    \hline
    Grit \# & $S_a$ ($\mu m$)   & Area Size ($\mu m^2$) \\ \hline
    100     & 106.24            & 198030.41    \\ \hline
    1000    & 14.36             & 198416.02    \\ \hline
    5000    & 1.77              & 153306.92    \\ \hline
    10000   & 0.94              & 149406.55    \\ \hline
    \end{tabular}
    \label{tbl:Sa}
\end{table}

\begin{figure}
    \centering
    \includegraphics[width=1\linewidth]{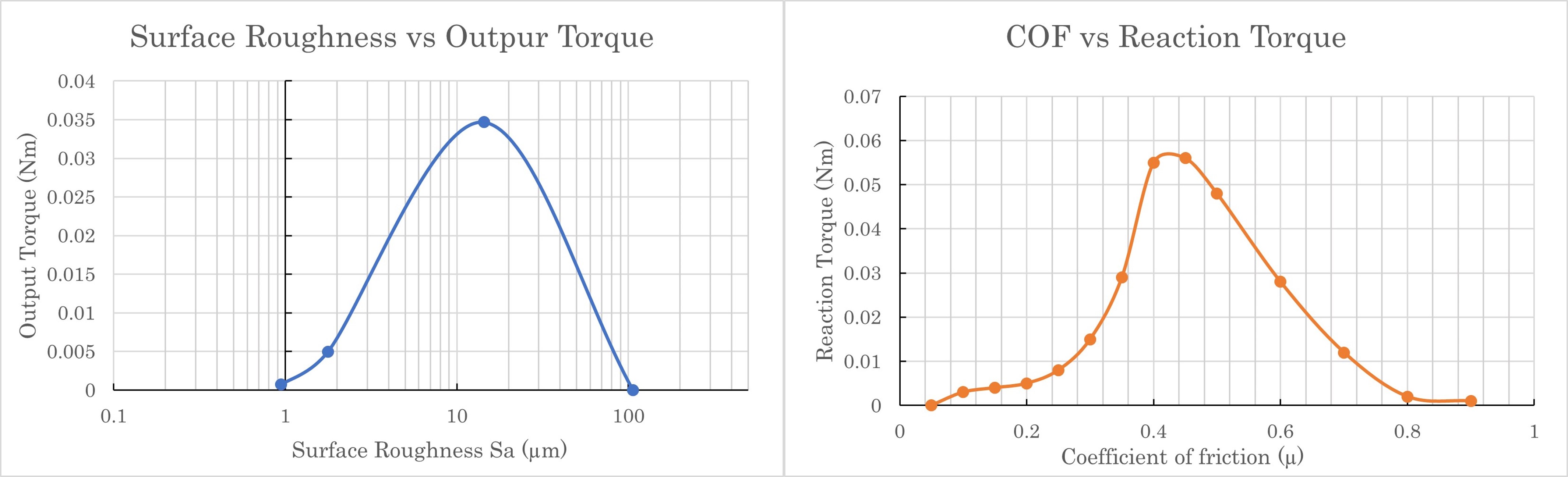}
    \caption{The surface roughness versus output torque experimental results compare with the COF versus reaction torque simulation results. The trends of both images are similar.}
    \label{fig:cofra}
\end{figure}

Based on the surface roughness value we measured for each stator, we chose the highest output torque of each stator performed measured in our previous work \cite{zhao2024design}, we can get the surface roughness versus output torque experimental results, which are shown in Figure \ref{fig:cofra} left figure. Results indicate that the surface roughness affected the torque performance significantly, too small or too large $S_a$ reduced the torque output value. If $S_a$ was too small, over smooth contacting surface will cause rotor slip. On the contrary, too large $S_a$ value indicates a rough surface with large surface geometry including spike and notch, and they may be larger than the surface traveling wave amplitude generated from the stator. In \cite{zhao2021preliminary,zhao2024study} we observed the amplitude of the wave can reach up to 300nm with a notched design, and the final motor design can reach even higher wave amplitude. However, the results from measurement in Table \ref{tbl:Sa} indicate that the rough surface can reach up to 106$\mu m$, which may be larger than the amplitude of the traveling wave, and can stop the rotor from rotating. Figure \ref{fig:cofra} right figure shows the simulation results of the plastic stator from Figure \ref{fig:FCsim}, a similar conclusion is observed that a range of COF was suggested to perform large reaction torque. Although the COF simulation and the surface roughness effect are not directly proportional to each other, they are all related to friction generation, and they share the same trends with motor torque performance, which suggested a practical solution for motor design, and the simulation model can help to find the optimal COF range for better supporting the motor design work.

\section{Conclusion and Disscussion}

In this work, a preliminary FEM simulation of a stator-rotor coupling system using COMSOL Multiphysics was presented, then using this model we studied how the prepressure and coefficient of friction (COF) affected the torque, and finally validate with experiment to equivalent tendency. The simulation model created was one of the limited models focusing on the motion of a whole motor, which can be a platform for further modeling studies. 

The prepressure influence was studied with this model. By validating with the physical experiment, a 40\% overshoot was observed between the simulation and experiment. Furthermore, the parameters and configuration used in the current stage simulation may still not be accurate and appropriate. Further modification with modeling including adding constraints to the rotor, changing prepressure applying the configuration, and adding new physical nodes can be considered. The tolerance from manual fabrication and COF value might cause the misalignment. For COF simulation, it is obvious that the friction coefficient is not a constant value at different prepressure. It indicates that the contact area increases with an increase in the prepressure due to the tribomaterial's toughness. Because the surface roughness is not related to the COF directly, in this study only the trends from the COF simulation and surface roughness were compared.

In this study, the thermal expansion and temperature interference are not discussed in the simulation, however, the temperature will significantly influence the surface property and further affect the friction condition. Furthermore, the thermal expansion will also affect the material propriety and cause some change in the motor operating status. A temperature-based study should be added to the simulation for accurate modeling. Moreover, cluster computing should also be involved considering the complex non-linear model always needs weeks to finish the calculation. By utilizing the cluster computing method, reduced time is excepted and can add more physics nodes, such as thermal expansion and heat transfer for a more precise simulation.

In this work, we configured the rotor as a rigid body considering a realistic situation of a small area contacting from multiple wave amplitude peaks, however, the friction layer of Teflon is soft and performs large deformation when contacting with wave amplitude peaks generated from excited stator. Future work should investigate the simulation performance with the non-rigid body configuration of the rotor. Moreover, the reaction torque from the simulation is performing an oscillating result between positive and negative values. Realistically, the motor inertia is likely filtering this periodic torque and an integral of the curve after the motor reaches a steady state. Future study of how the periodic torque correlates to the output torque of the motor is required. 

The contact mechanism of USM is a greatly complex non-linear problem, which makes it very difficult to conclude an analytical model. This current simulation model needs further development and study, considering the nearly no convergence results and over a couple of days calculation time range. By adjusting the physics nodes and parameters to converge the model faster with faster calculation time, even using the developer UI design from COMSOL Multiphysics, this model can be developed into a universal testing and evaluating platform based on the FEM simulation, which gives the community a guidance and easy-study platform for motor parameters design and developing.

\bibliographystyle{IEEEtran}
\bibliography{Main}

\end{document}